\documentclass[twocolumn,final]{elsarticleclass}
\usepackage{graphicx} %add graph
\usepackage{subfigure} %
\usepackage{indentfirst} %indent
\usepackage{setspace}%
\usepackage{amssymb} % //R
\usepackage{amsmath} % //math newline
\usepackage{algorithm}
\usepackage{algorithmic}
\usepackage{subfigure}
\usepackage{multirow}% multirow
\usepackage[showframe=true]{}
\usepackage{changepage}
\biboptions{numbers,sort&compress}
\usepackage{lineno,hyperref}
\begin{document}

\begin{frontmatter}
\title{ Building Fast and Compact Convolutional Neural Networks for Offline Handwritten Chinese Character Recognition}

\author[mymainaddress]{Xuefeng Xiao}
\ead{xiaoxuefengchina@gmail.com}

\author[mymainaddress]{Lianwen Jin\corref{mycorrespondingauthor}}
\cortext[mycorrespondingauthor]{Corresponding author}
\ead{lianwen.jin@gmail.com}

\author[mymainaddress]{Yafeng Yang}
\author[mymainaddress]{Weixin Yang}
\author[mysecondaryaddress]{Jun Sun}
\author[mymainaddress]{Tianhai Chang}

\address[mymainaddress]{School of Electronic and Information Engineering, South China University of Technology, Guangzhou, China}
\address[mysecondaryaddress]{Fujitsu Research \& Development Center Co. Ltd., Beijing, China}

\begin{abstract}
Like other problems in computer vision, offline handwritten Chinese character recognition (HCCR) has achieved impressive results using convolutional neural network (CNN)-based methods. However, larger and deeper networks are needed to deliver state-of-the-art results in this domain. Such networks intuitively appear to incur high computational cost, and require the storage of a large number of parameters, which renders them unfeasible for deployment in portable devices. To solve this problem, we propose a Global Supervised Low-rank Expansion (GSLRE) method and an Adaptive Drop-weight (ADW) technique to solve the problems of speed and storage capacity.  We design a nine-layer CNN for HCCR consisting of 3,755 classes, and devise an algorithm that can reduce the network's computational cost by nine times and compress the network to 1/18 of the original size of the baseline model, with only a 0.21\% drop in accuracy. In tests, the proposed algorithm surpassed the best single-network performance reported thus far in the literature while requiring only 2.3 MB for storage. Furthermore, when integrated with our effective forward implementation, the recognition of an offline character image took only 9.7 ms on a CPU. Compared with the state-of-the-art CNN model for HCCR, our approach is approximately 30 times faster, yet 10 times more cost efficient.
\end{abstract}


\end{frontmatter}

%\linenumbers
 \section{Introduction}
	Offline handwritten Chinese character recognition (HCCR) has been applied to a number of applications, such as for recognizing historical documents, mail sorting, transcription of handwritten notes, and so on. Offline HCCR has drawn the attention of many researchers for over half a century \cite{kimura1987modified, jin2000deformation,dai2007chinese,long2008building,liu2013online,zhang2017online}. In the last few years, a number of traditional offline approaches have been proposed to improve HCCR performance but have yielded scant progress; the modified quadratic discriminant function (MQDF) \cite{kimura1987modified, liu2013online,yin2013icdar}-based methods are exemplary. There is hence a recognition in the literature that even the best traditional methods are far from mimicking human performance in this domain \cite{yin2013icdar}. Due to the availability of better computational hardware and massive amounts of training data in recent years, convolutional neural networks (CNNs), proposed by LeCun in the 1990s \cite{le1990handwritten,lecun1998gradient}, have been used to attain state-of-the-art performance in character recognition \cite{zhang2017online}. The multi-column deep neural network (MCDNN) \cite{cirecsan2013multi}, composed of several CNNs, was the first CNN used for HCCR. Zhang et al. \cite{zhang2017online} recently reported an accuracy of 96.95\% for recognition by extracting the traditional normalization-cooperated direction-decomposed feature map as input with a CNN. However, the computational cost and storage requirements still prevent the use of CNNs in portable devices, where power consumption and storage capacity are the major challenges.

\par Many researchers have tried to build fast and compact networks. In this vein, low-rank expansion \cite{denton2014exploiting,jaderberg2014speeding,zhang2015efficient,zhang2015accelerating} aims to reduce computational cost by decomposing the convolutional layer. According to \cite{han2015learning,guo2016dynamic}, network pruning is the most effective way to compress the CNN; it eliminates the redundant connections in each layer, following which weight quantization and Huffman encoding are applied to further reduce storage capacity. Although \cite{denton2014exploiting,jaderberg2014speeding,zhang2015efficient,zhang2015accelerating,han2015learning,guo2016dynamic} achieved impressive performance in accelerating and compressing the network, only a few studies have combined these methods to address the dual challenge of speed and storage capacity. Furthermore, to the best of our knowledge, no study has investigated whether these methods are still feasible for large-scale handwritten Chinese character recognition involving more than 3,700 classes of characters.

\begin{figure*}[htbp]
  \centering
  \includegraphics[width=0.78\textwidth]{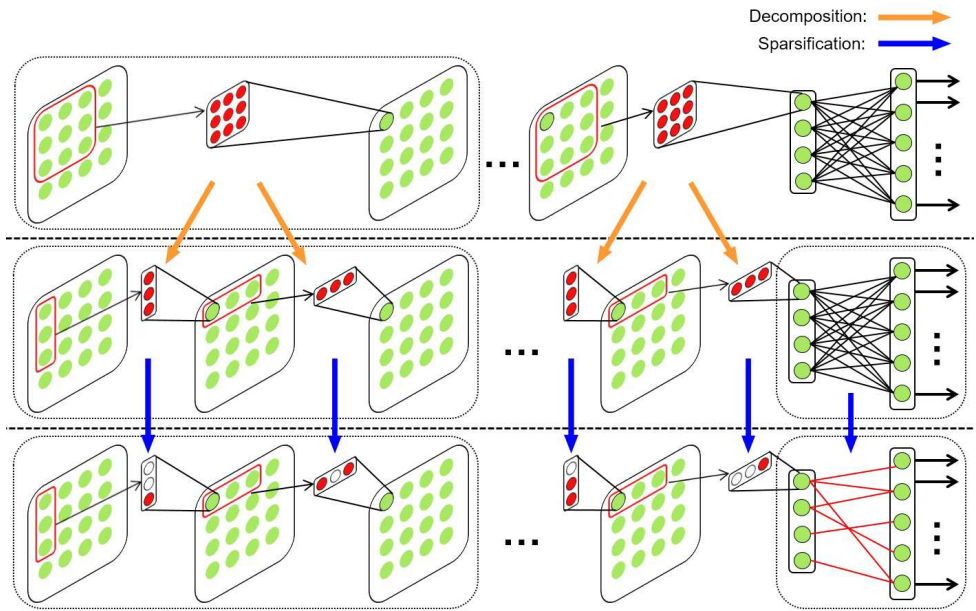}
  \caption{Overview of the proposed framework. The first step expands the convolutional layers, which involves decomposing each layer into two low-rank convolutional layers. The second step prunes redundant connections in the convolutional layers and the fully connected layers.}
  \label{total_struct}
\end{figure*}

\par In this paper, we propose a method to build a fast and compact CNN-based HCCR classifier. The method is shown in Fig. \ref{total_struct}; it unifies the advantages of low-rank expansion and network pruning. The first part employs low-rank expansion to decompose the convolutional layer for acceleration purposes, which renders the CNN deeper but compact. The motivation underlying the second part is to remove redundant connections in each layer's to further reduce the storage allocated to parameters and, hence, the computational cost of the entire network. However, in a previous study \cite{han2015learning,guo2016dynamic} on network pruning, the authors used a fixed threshold to prune the connections of each layer. Instead, we propose an Adaptive Drop-weight (ADW) technique that dynamically increases the threshold and gradually prunes out the parameters of each layer. Here, it comes another problem of the pruning ratio of each layer in previous work \cite{han2015learning,guo2016dynamic}, which may require numerous attempts for the determination of a suitable threshold for each layer in the context of a trade-off between accuracy drop and compression ratio, especially for some deep networks. Hence, to better address this problem, we propose Connection Redundancy Analysis (CRA) that can analyze redundancy in the connections of each layer and help maximize the pruning ratio of each layer with a tolerable reduction in accuracy.

\par In experiments involving offline HCCR, the proposed framework reduced by nine times the computational cost of, and by 18 times the parameter storage needed for, the designed CNN; and it degraded accuracy only by 0.21\%, which still surpassed the results for the best single-network CNN, reported in the literature thus far, on the ICDAR 2013 Offline HCCR Competition database. The network required only 2.3 MB of storage and took only 9.7 ms to process an offline character image on a single-thread CPU. Moreover, in order to further boost performance, we can increase the width and depth of the networks, or use a new CNN model such as GoogLeNet \cite{szegedy2015going} or Deep ResNet \cite{He_2016_CVPR}. This may help finally obtain new benchmarks for offline HCCR, but this is not our main concern in this paper.

\par The remainder of this paper is organized as follows: Section 2 reviews related work, and Section 3 elaborates on the architecture of the baseline network of the CNN used in our system. Section 4 introduces the adaptive drop-weight technique, whereas Section 5 details the connection redundancy analysis method. Section 6 describes global supervised low-rank expansions in detail, and Section 7 presents the experimental results, which include run time, parameter storage, and accuracy. The conclusions of this study and our future work are summarized in Section 8.

\section{Related Work}

\subsection{Offline HCCR}

\par Due to the success of CNNs, MQDF-based methods for offline HCCR have already reached their limit. The multi-column deep neural network (MCDNN) \cite{cirecsan2013multi}, consisting of several CNNs, was the first CNN used for offline HCCR. In an offline HCCR competition subsequently organized by ICDAR in 2013 \cite{yin2013icdar}, the method developed by the team from Fujitsu's R\&D Center won with an accuracy of 94.77\%. In 2014, they improved accuracy to 96.06\% by voting on four alternately trained relaxation convolutional neural networks (ATR-CNN) \cite{wu2014handwritten}. Zhong et al. \cite{zhong2015high} subsequently proposed combining traditional Gabor features with offline Chinese character images as network inputs, and used a streamlined version of GoogLeNet called HCCR-Gabor-GoogLeNet. They reported an accuracy of 96.35\%, and then that of 96.74\% for ensembling ten model and become the first one beyond human performance. The framework proposed by Zhou et al. \cite{zhou2015exploiting} is based on HCCR-GoogLeNet \cite{zhong2015high}; they used a Kronecker fully connected (KFC) layer to replace the layers after the four inception groups, and then followed by two fully connected layers, finally obtaining an accuracy of 96.63\%. Zhang et al. \cite{zhang2017online} recently combined traditional normalization-cooperated direction-decomposed feature maps and CNNs to obtain accuracy values of 96.95\% and 97.12\% by voting on three models.

\subsection{Accelerating and Compressing}

\par Most CNN structures, such as VGGNet \cite{Simonyan2014VeryDC}, AlexNet \cite{krizhevsky2012imagenet}, CaffeNet \cite{jia2014caffe}, and GoogLeNet \cite{szegedy2015going}, have similar properties: for example, the convolutional layers incur most of the computational cost and the fully connected layers contain the most network parameters. Despite the different potential avenues, existing approaches mainly concentrate on accelerating the convolutional layers and compressing the fully connected ones.

\begin{figure*}[htbp]
  \centering
  \includegraphics[width=0.99\textwidth]{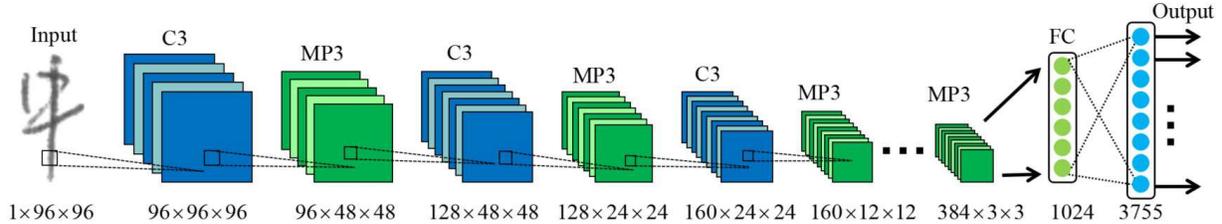}
  \caption{Architecture of the CNN for offline HCCR}
  \label{fig_network}
\end{figure*}

\par To reducing the computational cost of the convolutional layers, Cong and Xiao \cite{cong2014minimizing} used the Strassen algorithm for fast matrix multiplication to reduce the arithmetic complexity of the convolutional layer without loss in accuracy. Mathieu et al. \cite{mathieu2013fast} adopted the fast Fourier transform (FFTs) to convert convolutional calculations into pointwise products in the frequency domain for fast computation. Lavin et al. \cite{lavin2015fast} proposed using Winograd's minimal filtering algorithms to reduce the multiplication in the convolutional layers. Wu et al. \cite{wu2016quantized} recently proposed quantized convolutional neural networks that quantize the weights and transform computations into inner products in the convolutional layer. Nevertheless, computations of the convolutional layer are transformed into matrix multiplication by using the im2col algorithm \cite{Yanai2O16} and the BLAS (Basic Linear Algebra Subprograms) library. These tools are useful for faster CPU implementation of CNNs and cannot be used with the previously proposed method in \cite{cong2014minimizing,mathieu2013fast,lavin2015fast,wu2016quantized}. In this paper, we use the low-rank expansion-based method that can combine the matrix multiplication method by using the BLAS library. Jaderberg et al. \cite{jaderberg2014speeding} exploited the cross-channel or filter redundancy to formulate a low-rank basis for filters, and proposed filter and data reconstruction techniques for optimization. Zhang et al. \cite{zhang2015accelerating} improved their work by considering a non-linear case and asymmetric reconstruction for multiple layers to mitigate reconstruction error.

\par For fully connected layers, HashedNets proposed by Chen et al. \cite{chen2015compressing} uses the hash function to group weights into hash buckets, where connections in the same hash buckets share parameter value. Vanhoucke et al. \cite{vanhoucke2011improving} used an eight-bit fixed-point integer to replace the 32-bit floating point. Matthieu et al. \cite{courbariaux2016binarynet} proposed binarized neural networks that constrain weights and activations to +1 or -1, and replace most floating-point multiplications by one-bit exclusive-NOR operations. It is clear that this can reduce computational cost and parameter storage but, on the other hand, degrades network performance. Lin et al. \cite{Lin2016TowardsCN} used SVD-based low-rank expansions to compress the fully connected layers, and then used global error reconstruction to fine-tune the entire network. However, both these methods have low compression ratios, or seriously deteriorate network performance. Methods based on network pruning \cite{han2015learning} can significantly reduce parameter storage by learning important connections without compromising network performance. Deep compression was proposed by Han et al. \cite{Han2015DeepCC} to further reduce storage by combining network pruning, weight quantization, and Huffman coding. Guo et al. \cite{guo2016dynamic} proposed dynamic network surgery that can dynamically prune and splice connections based on Han's work \cite{han2015learning}.

\section{Architecture of Convolutional Neural Network}

\par As shown in Fig. \ref{fig_network}, we designed a nine-layer (only accounting for the convolutional layer and the fully connected layer) network for offline HCCR consisting of seven convolutional layers and two fully connected layers. Each of the first three convolutional layers are followed by a max-pooling layer. Following this, every two convolutional layers are followed by a max-pooling layer. The last max-pooling layer is followed by a fully connected layer, which contains 1,024 neurons. The last fully connected layer contains 3,755 neurons, and is used to perform the final classification. The overall architecture can be represented as Input-96C3-MP3-128C3-MP3-160C3-MP3-256C3-256C3-MP3-384C3-384C3-MP3-1024FC-Output.

\par We found that within a certain range, increasing the size of the input character image improved classification performance, but incurred higher computational cost. Hence, we fixed this effect of increasing size and computational cost by resizing the input characters into $96\times96$. In our baseline networks, all convolutional filters were $3\times3$, and a pixel was added to retain the size. Finally, the max-pooling operation was carried out over a $3\times3$ window with a stride of $2$.

\par In our proposed network, the parametric rectified linear unit (PReLU) \cite{He2015DelvingDI}, slightly different from the rectified linear unit (ReLU) \cite{Nair2010RectifiedLU}, was used to enable the network to easily converge and minimize the risk of overfitting to boost performance. Ioffe et al. \cite{ioffe2015batch} proposed batch normalization (BN), which can normalize nonlinear inputs and stabilize the distribution by reducing the internal covariate shift. It not only provides the liberty of using higher learning rates to expedite network convergence, but also ameliorates network performance at a negligible computational cost and storage. Moreover, for some deep networks, BN can effectively solve the problem of vanishing gradients. Therefore, all convolutional layers and the first fully connected layer were equipped with a BN layer, and the PReLU were added to each BN layer. Since the fully connected layers are quite redundant, we added the dropout \cite{srivastava2014dropout} layer between the two fully connected layers for regularization, where the ratio was set to 0.5.

\par The main difference between our proposed model and other available models for offline HCCR is that the former involves BN and PReLU in the network; hence, we refer to this baseline network as the HCCR-CNN9Layer. Although the CNN model used is quite simple, it is yielded state-of-the-art performance for HCCR.

\section{Adaptive Drop-Weight}

\begin{figure}[htbp]
  \centering
  \includegraphics[width=0.48\textwidth]{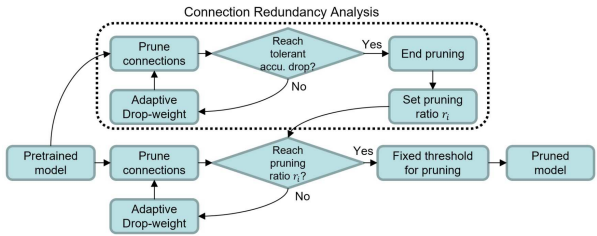}
  \caption{The pruning strategy}
  \label{adwcra}
\end{figure}

\par Our pruning scheme is shown in Fig. \ref{adwcra}; it consists of two parts. The first a new technique called Adaptive Drop-weight (ADW), which can gradually prune out the weighted connections of each layer by dynamically increasing the pruning threshold. When the pruning ratio reaches a value determined by the results of the CRA for the layer, we remember the threshold for further pruning.

\subsection{Pruning Threshold}

\par In the previous work on network pruning \cite{han2015learning,guo2016dynamic}, a fixed threshold was determined as follows:
\begin{equation}
\begin{aligned}
{{P_{th}} = \frac{\alpha }{N}\sum\limits_{i = 1}^N {\left| {{w_i}} \right|}  + \beta \sqrt {\frac{1}{N}\sum\limits_{i = 1}^N {(w{}_i - } \frac{1}{N}\sum\limits_{i = 1}^N {w{}_i} {)^2}}  + \lambda }.
\end{aligned}
\end{equation}

\par The pruning threshold ${P_{th}}$ depends on the weight of the layer ${w_i}$ in terms of average absolute value and variance. In order to render ${P_{th}}$ suitable for each layer, parameters $ \alpha,\beta,\lambda$ are selected by their empirical rules \cite{han2015learning,guo2016dynamic}. However, if the fixed threshold is too high, it leads to the pruning of a large number of connections at the outset, which results in drastic drop in performance. On the contrary, if the fixed threshold is too low, the compression ratio may be far from the desired value. In order to solve this problem, we propose using a dynamically increased threshold that gradually prunes out the weighted connections of each layer. This methodology can gradually lead the network toward self-adaptive pruning connections.

\subsection{Pruning Training Scheme}

\par In order to gradually prune redundant connections from each layer, we prune connections after every $I$ iterations (in experiments, we set $I = 10$). If we intend to prune the ratio ${r_i}$ in each layer that contains ${N_i}$ weights within ${T_1}$ iterations, the pruned number $p_i$ is increased by ${r_i}{N_i}I/{T_1}$ in each pruning iteration. The threshold is also gradually increased. During the iterations without the pruning process, the weights are updated with a gradient, and the pruned weights never come back. Once the desired pruning ratio is reached, the increasing threshold is fixed and noted for further pruning of the layer until pruning ends after ${T_2}$ iterations. This pruning process has been described in detail in Algorithm \ref{ADW}.

\par In order to further compress the network and improve performance, we employ the strategy proposed in \cite{Han2015DeepCC} to quantize weights. A k-means clustering algorithm is used to cluster weights for each layer of a pruned network. The quantized pruned network is then fine-tuned, which may result in better network performance.

\begin{algorithm}[htb]
\caption{ Adaptive Drop-weight for pruning the redundant connections.}
\label{ADW}
\begin{algorithmic}[1] %
\REQUIRE ~~\\ %
The weights in each layers $W_i$,the number of pruning layers $L$;
\ENSURE ~~\\ %
The sparse weights in each layer;
\WHILE{$t \le T_1$}
\IF{$t \% I = 0$}
\FOR{$i=1$ to $L$}
\STATE Only update the non-zero value in $W_i$
\STATE $p_i \Leftarrow p_i + {r_i}{N_i}I/{T_1}$\label{add_n}
\STATE Find the $p_i$-th smallest absolute value in $W_i$ to set $P_{th\_i}$
\STATE The absolute values of weights in $W_i$ below $P_{th\_i}$ are set to zero
\ENDFOR
\ELSE
\FOR{$i=1$ to $L$}
\STATE Only update the non-zero value in $W_i$
\ENDFOR
\ENDIF
\STATE $t \Leftarrow t + 1$
\ENDWHILE
\WHILE{$T_1 < t < T_2$}
\FOR{$i=1$ to $L$}
\STATE Only update the non-zero value in $W_i$
\STATE The absolute value of weights in $W_i$ below $P_{th\_i}$ are set to zero
\ENDFOR
\STATE $t \Leftarrow t + 1$
\ENDWHILE
\end{algorithmic}
\end{algorithm}

\section{Connection Redundancy Analysis}

\par The deep neural network consists of many layers, and each plays a significant role in the network. There are inevitably various redundancies in each layer, especially in the large gap between the convolutional layer and the fully connected layer. It makes sense that pruning ratio be determined by the redundant connections of each layers, and that the same pruning ratio not be applied to all layers. Nevertheless, retrospective work by \cite{han2015learning,guo2016dynamic} involved a fixed threshold ${P_{th}}$ based on the relevant layer's weights to prune the connection. Hence, numerous experiments are needed to find the pertinent values of $ \alpha,\beta,$ and $\lambda$ for the pruning threshold ${P_{th}}$ for each layer; as is self-evident, this is very time consuming, especially for deep networks.

\par In order to better address the above issue, we propose a Connection Redundancy Analysis (CRA) method that analyzes each layer's redundancy and can help us set a suitable value for the pruning ratio $r_i$ for it. Inspired by \cite{iandola2016squeezenet}, a sensitivity analysis was carried out to analyze the importance of a layer's parameters for network performance. Iandola et al. \cite{iandola2016squeezenet} implemented the strategy of directly pruning half the parameters with the smallest absolute values and carried out the experiment separately for each layer. After testing the pruned networks, network performance was examined. However, this strategy can only highlight the important parameters of a given layer with regard to performance, and cannot help determine how many connections are redundant.

\par To carry out the Connection Redundancy Analysis, we separately conducted an experiment for each layer. While carrying out the experiment on a layer, we fixed the weights of other layers and pruned only that layer. By using our proposed Adaptive Drop-weight as pruning strategy, we gradually pruned each layer's redundant connections, which were thought to gradually degrade network performance. When the drop in accuracy was beyond a given tolerance level, we knew how many connections had been pruned, which guided us in further pruning the network.

\par Since the proposed CRA was implemented to prune out the layers separately, it is difficult to analyze the scenario where all layers are pruned together. However, it may guide us in setting a proper pruning ratio for each layer. The ultimate goal of the CRA is to maximize the compression ratio under a tolerable reduction of accuracy rate, which desires further research.

\section{Grobal Supervised Low Rank Expansion}

\subsection{Decomposition Scheme}

\begin{figure}[htb]
\centering

\subfigure[]{
\label{original_conv}
\includegraphics[height=0.11\textwidth]{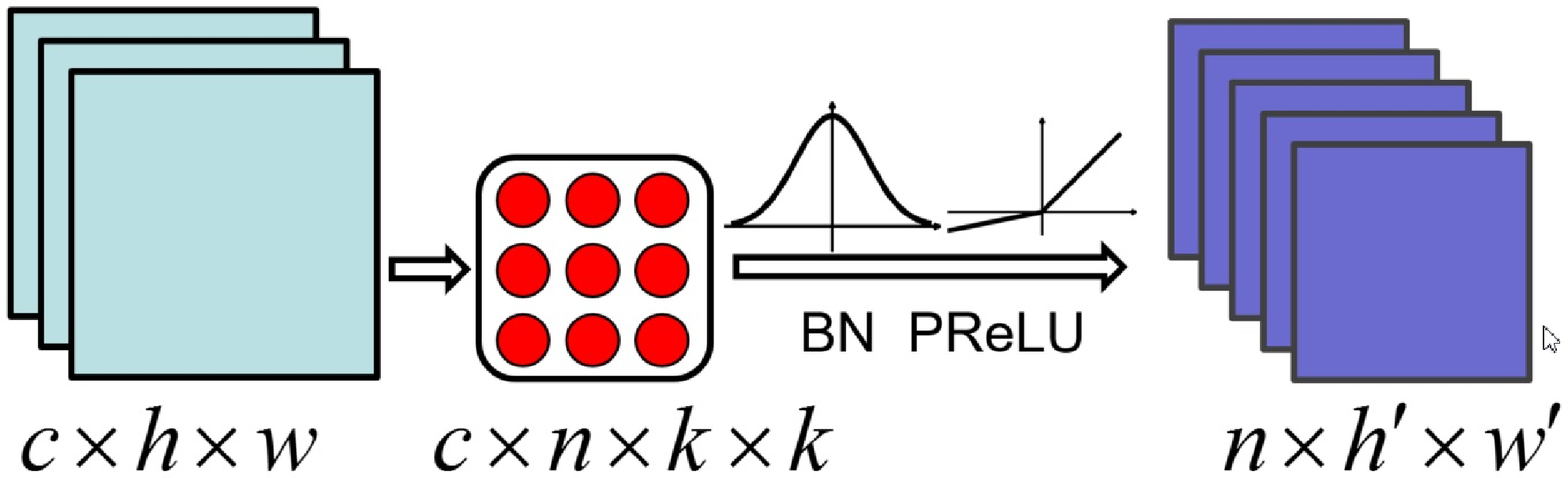}}

\subfigure[]{
\label{low_rank_conv}
\includegraphics[height=0.11\textwidth]{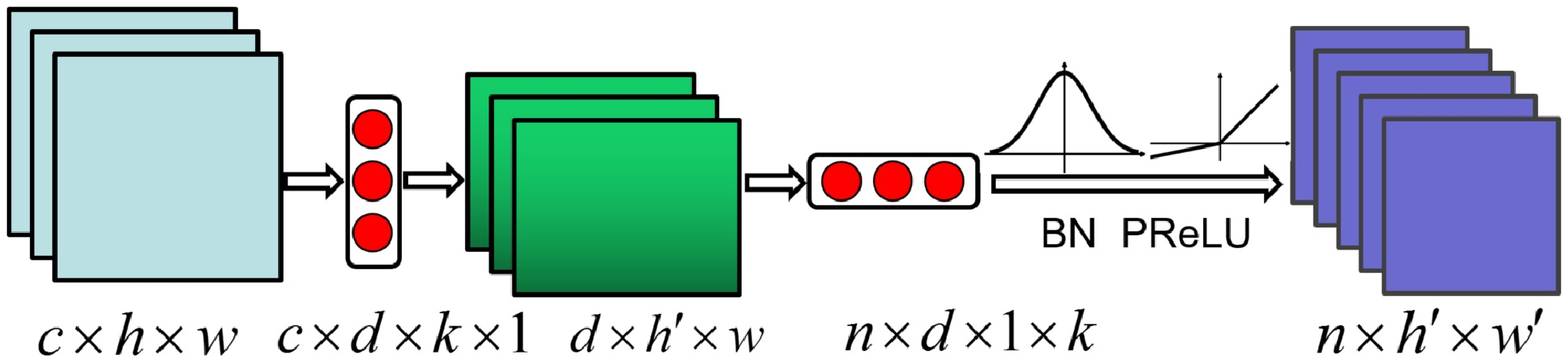}}

\caption{(a) The original convolutional layer with the BN and the PReLU layer. (b) The convolutional layer of low-rank expansion with the BN and the PReLU layer.}
\label{fig:2}
\end{figure}

\par For the original convolutional layer illustrated in Fig. \ref{original_conv}, the input feature map is a three-dimensional (3D) vector $X \in {\mathbb{R} ^{C \times H \times W}}$, where $C$ is the channel of the input feature map, and $H$ and $W$ are its height and width, respectively. The output feature map is also a 3D vector $Y \in {\mathbb{R} ^{N \times H' \times W'}}$, where $N$ is the channel of the output feature map, and $H'$ and $W'$ are its height and width, respectively. The kernel matrix is a 4D vector $W \in {\mathbb{R} ^{C \times N \times K \times K}}$, where the size of the kernel is $K \times K$. The output feature map can be calculated by
\begin{equation} \label{ECS}
\begin{aligned}
Y&(n,h',w') = \\
 & \sum\limits_{c = 1}^C {\sum\limits_{i = 1}^K {\sum\limits_{j = 1}^K {W(n,c,i,j)X(c,h' + i - 1,w' + j - 1)} } }.
\end{aligned}
\end{equation}
\par We know that the computational cost of the direct convolutional layer is $O(CN{K^2}H'W')$.

\par By carrying out the low-rank expansion shown in Fig. \ref{low_rank_conv}, that the input feature map originally convolved with the square filter, will be transformed into the input feature map convolved with two low rank filters. The first one is the input convolved with the vertical kernel $T \in {\mathbb{R} ^{C \times D \times K \times 1}}$, where $D$ is the output feature number the decomposed layer. The first output is
\begin{equation}
\begin{aligned}
M&(d,h',w) = \\
 & \sum\limits_{c = 1}^C {\sum\limits_{i = 1}^K {T(d,c,i,1)X(c,h' + i - 1,w)} },
\end{aligned}
\end{equation} where the computational cost by the first convolution is $O(CDKH'W)$.
\par Then the output $M \in {\mathbb{R} ^{D \times H' \times W}}$ convolves with horizontal kernel $V \in {\mathbb{R} ^{D \times N \times 1 \times K}}$, and the final output is calculated by
\begin{equation}
\begin{aligned}
Y&(n,h',w') =  \\
&\sum\limits_{d = 1}^D {\sum\limits_{j = 1}^K {V(n,d,1,j)M(d,h',w' + j - 1)} }.
\end{aligned}
\end{equation}
\par The computational cost by the second convolution is given by $O(NDKH'W')$. If the two low-rank expansions are considered together, the computational cost is  $O(DKH'(NW' + CW))$.
\par So if we want to accelerate $x$ time for the convolutional layer, $D$ can be determined as
\begin{equation}\label{midder_low_rank_value}
D = \frac{{CNKW'}}{{(CW + NW')x}}.
\end{equation}

\subsection{Training scheme}

\par
\begin{figure}[htb]
\centering

\subfigure[]{
\label{l2loss}
\includegraphics[width=0.45\textwidth]{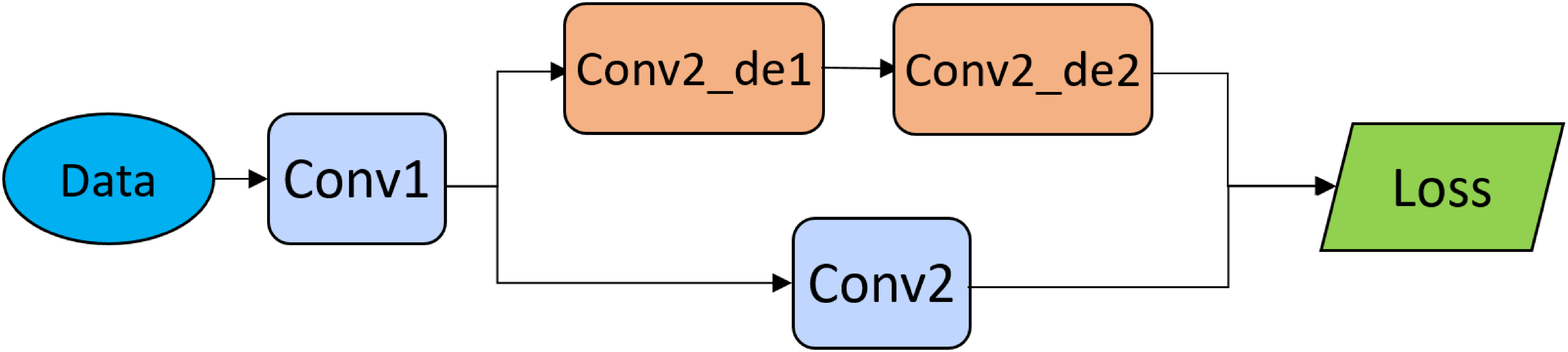}}

\subfigure[]{
\label{softmaxloss}
\includegraphics[width=0.45\textwidth]{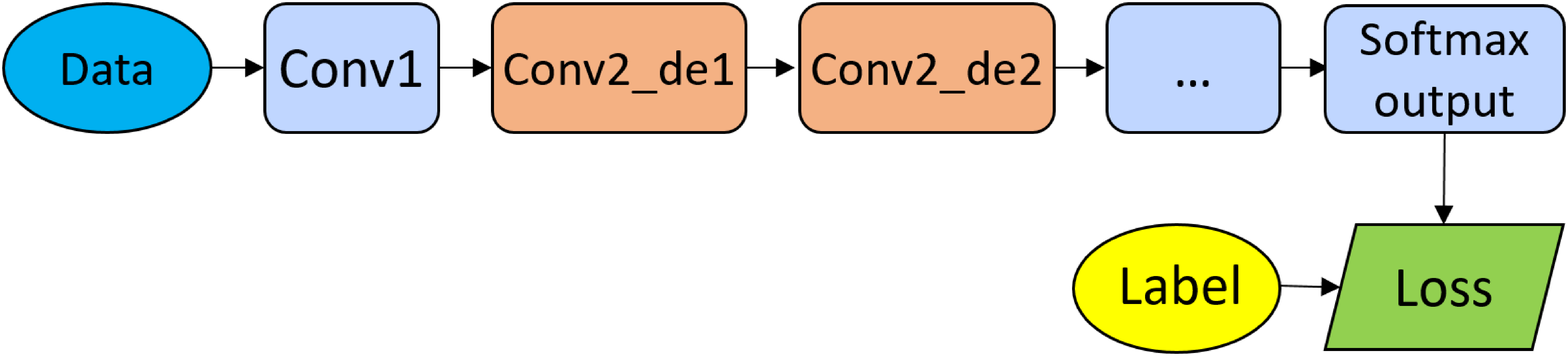}}

\caption{(a) The local output is used as the supervisor to train the low-rank layer. (b) The true label-guided global loss function is used as the supervisor to train the low-rank layer.}
\label{fig:2}
\end{figure}

 In past work \cite{jaderberg2014speeding,zhang2015efficient,zhang2015accelerating}, the output of each layer was used as a supervisor to learn the low-rank filter for that layer. This method was mainly devised to minimize the reconstruction error between the local output and the low-rank approximation output, as shown in Fig. \ref{l2loss}. We refer to this strategy as Local Supervised Low-rank Expansion (LSLRE). We think that while using the output of the local layer to guide the low-rank expansion is a reasonable and straightforward strategy, it does not a direct relationship with the performance of global classification.

\par Thus, we propose Global Supervised Low-rank Expansion (GSLRE), which uses the label as supervisor. The training scheme is shown in Fig. \ref{softmaxloss}. The training process is conducted in a layer-by-layer manner. For a specific layer, the original convolutional layer, say, the second layer Conv2 is decomposed into two smaller layers, Conv2\_de1 and Conv2\_de2  (see Fig. \ref{softmaxloss}). The parameters of Conv2\_de1 and Conv2\_de2 are determined through back-propagation using the SGD algorithm, based on the loss function for the entire network. It is worth mentioning that during the training of the specific convolutional layer, the parameters of other convolutional layers are kept fixed. This is because our network is equipped with the BN layer, which can enable gradients to smoothly pass into lower-level layers even though the network deepens.

\par Since the first convolutional layer is hard to approximate (as mentioned in \cite{denton2014exploiting,jaderberg2014speeding}), and plays an important role in extracting features from the original image, we begin our low-rank training scheme at the second convolutional layer. Once the second layer has been decomposed and trained adequately, we begin decomposing the third layer. When training the third convolutional layer, the parameters of both the second and the third layers are learned and updated according to the SGD-BP algorithm. In this way, all convolutional layers are decomposed and trained. Finally, since the first convolutional layer and the parameters of the fully connected layer are always fixed during the above low-rank expansion training process, we need to fine-tune the whole network to further improve overall performance.

\section{Experiment}

\par We evaluated our method on an offline HCCR task. The training data was taken from the CASIA-HWDB1.0 and CASIA-HWDB1.1 \cite{liu2011casia} databases, written by 300 and 420 people, respectively. The datasets contained 2,678,424 samples in total. The test data were datasets used in the ICDAR-2013 offline competition \cite{yin2013icdar}, which contained 224,419 samples written by 60 people. The training and testing databases were written by different people, and contained 3,755 classes.

\subsection{Baseline Model}

\par We trained our baseline network (Fig. \ref{fig_network}) on the Caffe \cite{jia2014caffe} deep learning platform. We used the method of mini-batch gradient descent with momentum for training. The mini-batch size was set to 128 and momentum to 0.9. In order to make training data independent of each mini-batch, we shuffled the data before training. Since our proposed network is equipped with a BN layer, it allowed us to use higher learning rates to accelerate converge. We hence initialize the learning rate at 0.1, and then reduced it $\times 0.1$ every 70,000 iterations. Training was completed after 300,000 iterations, and we obtained an accuracy of 97.30\%.

\par Fig. \ref{fig_network} shows our proposed network. We realized that the first convolutional layer plays an important role in extracting features from the original image \cite{denton2014exploiting,jaderberg2014speeding}, and incurs lower computational cost and smaller parameter storage (less than 1\% of the entire network). Thus, it is intuitive that the parameters of this layer should not be modified.

\subsection{The evaluation of the Global Supervised Low Rank Expansions}

\begin{table}[htbp]
\centering
\caption{The results of accelerating 4 times}
\begin{tabular}{l|c}
\hline
Method &  Accuracy \\
\hline
Baseline & 97.30\% \\
\hline
Direct Low Rank training   & 97.10\% \\
\hline
GSLRE &  97.27\% \\
\hline
\end{tabular}
\label{accelerating_result_table}
\end{table}

\begin{table*}[htbp]
 \begin{adjustwidth}{-0.9cm}{}
\centering
\caption{The compression results on LeNet-5 and LeNet-300-100.}
\begin{tabular}{l|l|l|l|l|l|l|l}
\hline
Network & Layer & Params. &Params. after  & Params. after& Params. after & Pruning+ & Pruning+ \\
& & &pruning\cite{han2015learning} &pruning\cite{guo2016dynamic} &pruning ours&clustering\cite{Han2015DeepCC} & clustering ours\\
\hline
\hline
\multirow{7}{*}{LeNet-5} &conv1  & 0.5K & 66\% & 14.2\% & 27.2\% & 78.5\% & 10.8\%\\
\cline{2-8}
&conv2  & 25K & 12\% & 3.1\% & 3.7\% & 6.0\% & 1.5\%\\
\cline{2-8}
&fc1  & 400K & 8\% & 0.7\% & 0.4\% & 2.7\% & 0.28\%\\
\cline{2-8}
&fc2  & 5K & 19\% & 4.3\% & 8.8\% & 6.9\% & 3.0\%\\
\cline{2-8}
&total & 431k & 8\%(12X) & 0.9\%(108X) &\textbf{0.75\%(133X)} & 3.05\%(33X) & \textbf{0.4\%(250X)}\\
\cline{2-8}
& \multicolumn{2}{|l|}{Baseline Accu.} & 99.20\% & 99.09\% & 99.11\% & 99.20\% & 99.11\%\\
\cline{2-8}
& \multicolumn{2}{|l|}{Pruned Accu.} & 99.23\% & 99.09\% & 99.11\% & 99.26\% & 99.12\%\\
\hline
\hline
\multirow{6}{*}{LeNet-300-100}
&fc1  & 236K & 8\% & 1.8\% & 1.5\% & 3.1\% & 0.8\%\\
\cline{2-8}
&fc2  & 30K & 9\% & 1.8\% & 2.8\% & 3.8\% & 1.5\%\\
\cline{2-8}
&fc3  & 1K & 26\% & 5.5\% & 8.5\% & 15.7\% & 5.8\%\\
\cline{2-8}
&total & 267k & 8\%(12X) & 1.8\%(56X) &\textbf{1.7\%(60X)} & 3.1\%(32X) & \textbf{0.9\%(113X)}\\
\cline{2-8}
& \multicolumn{2}{|l|}{Baseline Accu.}& 98.36\% & 97.72\% & 98.33\% & 98.36\% & 98.33\%\\
\cline{2-8}
& \multicolumn{2}{|l|}{Pruned Accu.} & 98.41\% & 98.01\% & 98.34\% & 98.42\% & 98.35\%\\
\hline
\end{tabular}
\label{tab:mnist_database}
 \end{adjustwidth}
\end{table*}

\par By using our proposed architecture, the baseline network was accelerated fourfold. Using Eq. \ref{midder_low_rank_value}, we calculated the number of feature maps each convolutional layer after decomposing it. In Table \ref{accelerating_result_table}, it is clear that we were able to accelerate the network fourfold with a negligible drop in accuracy by integrating our devised Global Supervised Low-rank Expansion training scheme. It was also shown that our decomposing training scheme can obtain better results than the direct training of a decomposed network architecture.

\subsection{The evaluation of the Adaptive Drop-weight}

\par We first applied our method to the MNIST database with the LeNet-300-100 and the LeNet-5 networks \cite{lecun1998gradient}. The MNIST dataset was designed for character recognition of handwritten digits. LeNet-5 is a convolutional network that contains two convolutional layers and two fully connected layers. LeNet-300-100 is a fully connected network with two hidden layers. The baseline models were trained on the Caffe \cite{jia2014caffe} deep learning platform without any data augmentation. We directly trained the LeNet-5 using the settings for the training parameters provided by Caffe. In this way, an accuracy of 99.11\% was obtained after training for 10,000 iterations. The training parameter settings of LeNet-300-100 were nearly identical to those for LeNet-5, and yielded an accuracy of 98.33\%.

\begin{table}[htbp]
\caption{The compression results on ICDAR 2013 dataset.}
\centering
\begin{tabular}{l|c|c|c}
\hline
Layer  & Params. & Params. after & After pruning\\
 & & pruning& + clustering \\
\hline
conv2\_de1  & 12K & 36.9\% & 11.3\%\\
\hline
conv2\_de2  & 15K & 42.7\% & 12.5\%\\
\hline
conv3\_de1  & 20K & 33.6\% & 10.2\%\\
\hline
conv3\_de2  & 24K & 41.9\% & 12.1\%\\
\hline
conv4\_1\_de1  & 33K & 35.3\% & 10.4\%\\
\hline
conv4\_1\_de2  & 53K & 37.6\% & 11.0\%\\
\hline
conv4\_2\_de1  & 68K & 39.6\% & 11.5\%\\
\hline
conv4\_2\_de2  & 68K & 39.1\% & 11.3\%\\
\hline
conv5\_1\_de1  & 78K & 40.0\% & 11.6\%\\
\hline
conv5\_1\_de2  & 118K & 35.3\% & 10.3\%\\
\hline
conv5\_2\_de1  & 142K & 33.9\% & 10.0\%\\
\hline
conv5\_2\_de2  & 142K & 37.0\% & 10.7\%\\
\hline
fc1  & 3.5M & 14.2\% & 4.4\%\\
\hline
fc2  & 3.8M & 35.1\% & 9.06\%\\
\hline
total  & 8.17M & 26.2\%(3.8x) & 7.2\%(13.9X)\\
\hline
\end{tabular}
\label{tab:offline_hccr_pruning}
\end{table}

\par As shown in Table \ref{tab:mnist_database}, with our proposed pruning strategy, we compressed LeNet-5 by a factor of 133 and LeNet-300-100 by that of 60 using the proposed ADW method. It surpassed the results in \cite{han2015learning}, which was the first application of network pruning for compression. Compared with recent work \cite{guo2016dynamic}, we achieved a higher pruning ratio and better accuracy, especially for LeNet-300-100. We also combined the methods of quantizing weights for further compression. Finally, we obtained a state-of-the-art compression ratio of 250 times for LeNet-5 and 113 times for LeNet-300-100 without any loss in accuracy.

\par Following this, we applied our proposed compression framework on an offline HCCR network that was accelerated fourfold, and contained 13 convolutional layers and two fully connected layers. In Table \ref{tab:offline_hccr_pruning}, it is evident that the entire network was compressed to approximately a quarter of its size using only the proposed ADW method. When we integrated the weight quantization, storage was further reduced approximately 14-fold with a drop of only 0.18\% in accuracy.

\par While pruning LeNet-5, we noticed that separately pruning the convolutional and the fully connected layers was a better choice to deal with the vanishing gradient problem than pruning these layers together, which was the strategy used in past work \cite{han2015learning,guo2016dynamic}. However, since our accelerated network was equipped with the BN layer, it enabled the gradient to smoothly pass in both forward and backward propagations, as also demonstrated in \cite{He_2016_CVPR}. In our experiment, we were able to prune the convolutional layers and fully connected layers together. This not only reduced training time for the pruning process, but also yielded higher accuracy and compression ratio at the same time.

\subsection{The evaluation of the Connection Redundancy Analysis}

\par We implemented our CRA on each layer in LeNet-5 and LeNet-300-100. With our proposed Adaptive Drop-weight pruning strategy, we gradually pruned the connections of each layer.

\par
\begin{figure}

  \label{craMnistResult}
  \centering
  \subfigure[]{
    \label{fig:subfig:a} %% label for first subfigure
    \includegraphics[width=2.6in]{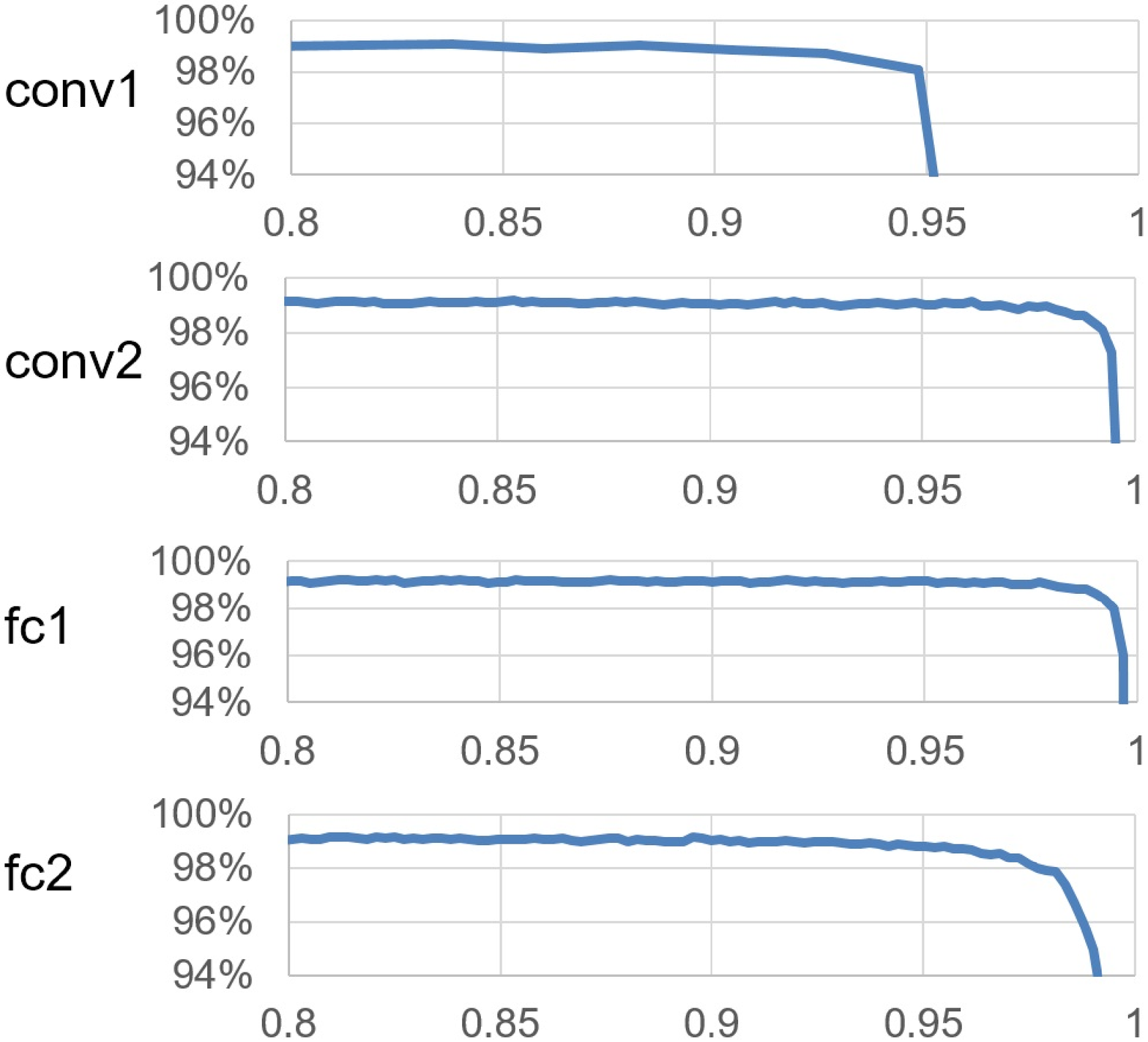}}
  \hspace{0.2in}
  \subfigure[]{
    \label{fig:subfig:b} %% label for second subfigure
    \includegraphics[width=2.6in]{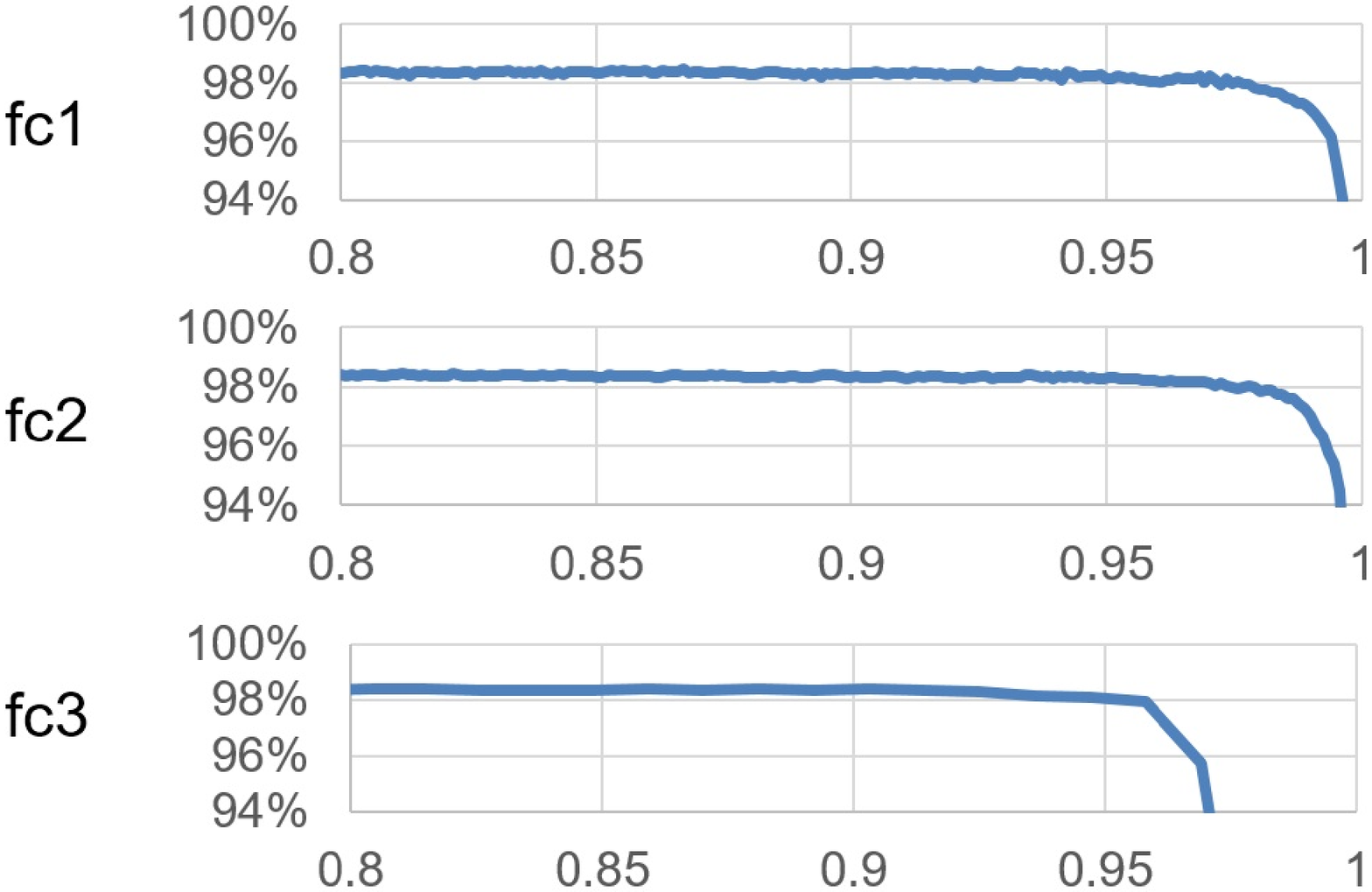}}
  \caption{The CRA results. The horizontal and vertical axes represent pruning ratio and network accuracy, respectively. (a) The results of gradually pruning each layer of LeNet-5. (b) The results of gradually pruning each layer of LeNet-300-100.}

\end{figure}

 In Fig. \ref{fig:subfig:a} and \ref{fig:subfig:b}, we see that at the start of the experiment, the network was less susceptible to the pruning ratio; but later on, drastically decreases down with the higher values of this pruning ratio. However, since each layer had different redundant connections, the pruning ratio of each was different. In our experiment, CRA was implemented with a tolerable accuracy drop of 0.1\% for each layer. Then, the pruning ratio was used to guide the pruning of the network.

\par
\begin{figure}[htbp]
  \centering
  \includegraphics[width=0.5\textwidth]{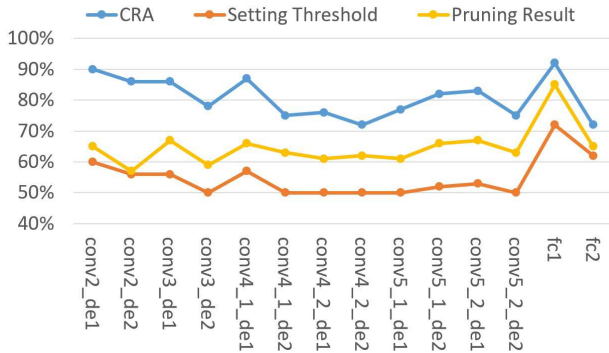}
  \caption{Connection Redundancy Analysis For HCCR-CNN9Layer-LR}
  \label{cra_analysis_hccr}
\end{figure}

 In the same way, CRA was applied to an offline HCCR network, which reduced computational cost fourfold in the convolutional layer. We then analyzed each layer's redundancy with a tolerable accuracy drop of 0.1\% to guide us in pruning. Since the convolutional layer had been accelerated four times, the redundancy therein was significantly eliminated. As shown in Fig. \ref{cra_analysis_hccr}, we set a much lower pruning ratio than the CRA results in convolutional layer to maintain accuracy at par.

\subsection{The results of accuracy}

\par Table \ref{tab:result_icdar} illustrates the results of different methods that achieved performance beyond human level on the ICDAR-2013 offline competition database as well as their network storage and FLOPs (multiply-adds) in detail.

\par We use our nine-layer network, shown in Fig. \ref{fig_network}, and achieved an accuracy of 97.30\%. We realized that the BN and the PReLU layers are quite effective for offline HCCR. Using our proposed GSLRE and ADW, we further reduced computational cost by nine times and parameter storage by 18 times with a only drop of 0.21\% in accuracy. This result still surpassed the best single-network performance in line 7 in Table \ref{tab:result_icdar}, but our model simultaneously involved considerably less parameter storage and incurred lower computational cost.

\begin{table*}[htbp]
 \begin{adjustwidth}{-0.001cm}{}
\caption{The results on ICDAR-2013 offline HCCR competition. }
\centering
\begin{tabular}{|c|l|c|c|c|c|}
\hline
No.&Method &Ref.& Storage(MB)  & FLOPs(${\times 10^8}$) & Top-1 Error(\%) \\
\hline
1 & Human Performance & \cite{yin2013icdar} & n/a & n/a &  3.87 \\
\hline
2 & HCCR-Gabor-GoogLeNet &\cite{zhong2015high} & 24.7 & 3.58 & 3.65  \\
\hline
3 & HCCR-GoogLeNet-Ensemble-10 &\cite{zhong2015high} & 247 & n/a & 3.26 \\
\hline
4 & CNN-Single  &\cite{chen2015beyond}& 190& 1.33  & 3.42 \\
\hline
5 & CNN-Ensemble-5  &\cite{chen2015beyond}& 950 & 6.65  & 3.21 \\
\hline
6 & Kronecker Fully-Connected(KFC) &\cite{zhou2015exploiting} & n/a & n/a & 3.37 \\
\hline
7 & DirectMap + ConvNet &\cite{zhang2017online} & 23.5 & 2.63 & \textbf{3.05} \\
\hline
8 & DirectMap + ConvNet + Ensemble-3 &\cite{zhang2017online} & 70.5 & 7.89 & 2.88 \\
\hline
\hline
9 &HCCR-CNN9Layer  & ours & 41.5 &  5.94 & 2.70  \\
\hline
10 & HCCR-CNN9Layer+GSLRE 4X & ours & 31.2 & 1.52 & 2.73 \\
\hline
11 & HCCR-CNN9Layer+GSLRE 4X +ADW & ours & \textbf{2.3(18X)} & \textbf{0.65(9X)} & \textbf{2.91}  \\
\hline
12 & HCCR-CNN12Layer & ours & 48.7 & 12 & 2.41  \\
\hline
13 & HCCR-CNN12Layer+GSLRE 4X & ours & 32.7 & 2.99 & 2.53  \\
\hline
14 & HCCR-CNN12Layer+GSLRE 4X+ADW & ours & 3.0 & 1.25 & 2.60  \\
\hline
\end{tabular}
\label{tab:result_icdar}
\end{adjustwidth}
\end{table*}

\par It was clear that the larger and deeper the network, the better it performed. Hence, based on our proposed network, we added more convolutional layers: Input-96C3-MP3-128C3-128C3-MP3-192C3-192C3-MP3-256C3-256C3-MP3-384C3-384C3-384C3-MP3-1024FC-Output. We refer to this large network as HCCR-CNN12Layer, which yielded an accuracy of 97.59\%, as shown on line 12 in Table \ref{tab:result_icdar}. Then, combining our GSLRE and ADW, we were still able to reduce computational cost 16-fold and parameter storage 10-fold with only a drop of 0.19\% in accuracy.

\subsection{The results of the forward implementation}

\par The run time of the network is crucial for applying offline HCCR to deal with real-time tasks. Other techniques can be deployed for accelerating CNNs for real-time applications. Loop unrolling (LU) is a well-known and efficient strategy to improve speed, especially for large loops. Using the im2col algorithm, convolutional computations were converted into matrix-matrix multiplication using the BLAS library\footnote{In the following experiments, we used Intel MKL as the BLAS library, available at https://software.intel.com/en-us/intel-mkl.}, which has been shown to be an efficient way for CPU-based implementation of CNNs. By using the BLAS library, the fully connected layers were directly implemented into matrix-vector multiplication. Moreover, when we eliminated connections in each layer using our proposed ADW method, we used sparse matrix-matrix multiplication and sparse matrix-vector multiplication, respectively, for the convolutional layer and the fully connected layer. However, we found that if the layer was not sparse enough, performance degraded. In our proposed network, we simply applied sparse matrix-vector multiplication only to compute the fully connected layer.

\begin{table}[htbp]
\caption{The runing time for processing one character on a CPU.}
\centering
\begin{tabular}{l|c|c|c}
\hline
Method  &Storage & Accuracy & Time \\
\hline
Direct Calculation & 41.5MB & 97.30\%  & 1368ms\\
\hline
LU  & 41.5MB & 97.30\%  & 492ms\\
\hline
LU+4X & 31.2MB & 97.27\% & 118ms \\
\hline
LU+BLAS  & 41.5MB & 97.30\%  & 21.1ms\\
\hline
LU+4X+BLAS & 31.2MB & 97.27\% & 10.1ms \\
\hline
LU+4X+Sparse+BLAS  & 2.3MB & 97.09\% & 9.7ms \\
\hline
DirectMap+ConvNet\cite{zhang2017online}& 23.5MB & 96.95\% & 296.9ms \\
\hline
\end{tabular}
\label{tab:runing_time}
\end{table}

\par We compared the forward run time with different strategies on a single-threaded CPU. The experiments were carried out on a single desktop PC equipped with 3.60 GHz Intel Core i7-6700 and 16 GB of memory. From Table \ref{tab:runing_time}, we see that when we did not use a technique to accelerate the CNN, the run time was long (1369 ms per character). When we simply used loop unrolling for all layers, the run time was reduced. When we used our acceleration method and reduced computational cost fourfold, the run time was also reduced approximately fourfold (from 492 ms to 118 ms). Then, all convolutional layers and the fully connected layer were computed by adopting matrix-matrix and matrix-vector multiplication, respectively, with the BLAS library. Loop unrolling was also applied to all other layers. The run time decreased significantly. Finally, using our compression method to prune redundant connections in the convolutional layers and the fully connected layers, we employed sparse matrix-vector multiplication to implement the computations in the fully connected layer. In this way, we achieved a fast and compact CNN model for large-scale HCCR with a speed of 9.7 ms/char but only 2.3 of MB storage. We compared the forward run time implemented by Zhang et al. \cite{zhang2017online}(the last row) with that of our model. The proposed forward implementation method was clearly more effective than Zhang's \cite{zhang2017online} method: it was approximately 30 times faster but 10 times smaller. The source code of our fast and compact CNN model's forword implementation will soon be made publicly available.

\section{Conclusion}

\par In this paper, we proposed an effective approach for accelerating and compressing a CNN for large-scale HCCR involving 3,755 classes of Chinese characters. We proposed a Global Supervised Low-rank Expansion to accelerate calculations in the convolutional layers, and an Adaptive Drop-weight method to remove redundant connections by using a dynamic increase in the pruning threshold of each layer. We also proposed Connection Redundancy Analysis technology to analyze redundant connections in each layer in order to guide the pruning of the CNN without compromising the performance of the network.

\par In future work, we plan to apply the proposed framework to other fields, such as image classification and object detection. These ideas can also be used to address deep recurrent neural networks \cite{Graves2012Supervised}, especially for long short-term memory, as they are viable deep-learning models to deal with such time sequence-based problems as online handwritten character/text recognition \cite{zhang2016drawing,xie2016learning}.

%\bibliography{reference_file}

\bibliographystyle{model2-names}

\end{document}